\documentclass[conference]{IEEEtran}
\IEEEoverridecommandlockouts
\usepackage{url}

\usepackage{breakurl}
\usepackage[breaklinks]{hyperref}
\usepackage{tabularx}
\usepackage{multirow}
\usepackage{amsmath}
\usepackage{caption}
\usepackage{subfig}
\usepackage{tikz}
  \usetikzlibrary{positioning}
  \usetikzlibrary{tikzmark} % arrows in tex
  \usetikzlibrary{arrows}   % arrows in tex
  \usetikzlibrary{calc}     % (node)+(3cm,2cm)
  \tikzstyle{every picture}+=[remember picture]
\newcommand\norm[1]{\left\lVert#1\right\rVert}

\makeatletter
\def\endthebibliography{%
	\def\@noitemerr{\@latex@warning{Empty `thebibliography' environment}}%
	\endlist
}
\makeatother

\usepackage{cite}
\usepackage{amsmath,amssymb,amsfonts}
\usepackage{algorithmic}
\usepackage{booktabs}
\usepackage{graphicx}
\usepackage{subcaption}
\usepackage{textcomp}
\usepackage{xcolor}
\usepackage[left=0.75in,right=0.75in,top=1in,bottom=0.75in]{geometry}
\def\BibTeX{{\rm B\kern-.05em{\sc i\kern-.025em b}\kern-.08em
    T\kern-.1667em\lower.7ex\hbox{E}\kern-.125emX}}
\begin{document}

\title{\LARGE \bf Augmenting Safety-Critical Driving Scenarios while \\ Preserving Similarity to Expert Trajectories
}

% \author{\IEEEauthorblockN{Hamidreza Mirkhani}
% \IEEEauthorblockA{\textit{Noah's Arc Laboratory} \\
% \textit{Huawei Technologies Canada}\\
% Hamidreza.Mirkhani@Huawei.com}
% \and
% \IEEEauthorblockN{Behzad Khamidehi}
% \IEEEauthorblockA{\textit{Noah's Arc Laboratory} \\
% \textit{Huawei Technologies Canada}\\
% Behzad.Khamidehi@Huawei.com}
% \and
% \IEEEauthorblockN{Kasra Rezaee}
% \IEEEauthorblockA{\textit{Noah's Arc Laboratory} \\
% \textit{Huawei Technologies Canada}\\
% Kasra.Rezaei@Huawei.com}
% \and
% }

\author{
Hamidreza Mirkhani, Behzad Khamidehi, and Kasra Rezaee\\
{\textit{Noah's Ark Lab, Huawei Technologies Canada}} \\
{\textit{Emails: firstname.lastname@huawei.com}} \\%
%\normalsize $^{*}$These authors contributed equally \\% <-this % stops a space
}

\maketitle

\begin{abstract}
Trajectory augmentation serves as a means to mitigate distributional shift in imitation learning. However, imitating trajectories that inadequately represent the original expert data can result in undesirable behaviors, particularly in safety-critical scenarios. We propose a trajectory augmentation method designed to maintain similarity with expert trajectory data. To accomplish this, we first cluster trajectories to identify minority yet safety-critical groups. Then, we combine the trajectories within the same cluster through geometrical transformation to create new trajectories. These trajectories are then added to the training dataset, provided that they meet our specified safety-related criteria. Our experiments exhibit that training an imitation learning model using these augmented trajectories can significantly improve closed-loop performance.
\end{abstract}


\section{Introduction}
Data-driven planning models trained on high-quality human driving data can replicate the smooth and intelligent behaviors observed in those exemplary demonstrations. While this data can be collected on a large scale, their distribution often exhibits a high level of imbalance. This imbalance has the potential to induce a distributional shift in the trained models or prompt them to overfit to dominant scenarios, resulting in poor generalization \cite{distributional-shift-ross2011reduction, distributional-shift-pmlr-v9-ross10a}. Moreover, safety-related scenarios typically constitute minority clusters within the input data \cite{imitation-is-not-enough-lu2023imitation}. Neglecting these scenarios can have severe consequences, leading to catastrophic outcomes. A viable approach to tackle this problem is augmenting the data within minority yet crucial clusters, re-balancing the training datasets \cite{dataset-re-balancing-zhang2023deep}. Numerous studies have been conducted on data augmentation across various domains \cite{CV-Low-shot-Visual-Recognition-2018, CV-Low-shot-Δ-encoder-2018, CV-Interpolate-Faramarzi_2022, CV-Interpolate-pmlr-v97-verma19a, CV-interpolate-zhang2018mixup, DA-NLP-Attention-Score-yu2022tabas, DA-NLP-Pre-trained-model-kumar2021data, Liu_2018_CVPR}. To apply augmentation techniques in trajectory planning domain, it is necessary to establish a transformation function for trajectories. This function should preserve the primary characteristics of the original data within a complex environment. To address characteristics preservation, some efforts have been made to create traffic scenes through procedural models. That involves the introduction of dynamic agents based on a set of heuristic rules. The parameters of these heuristics are manually adjusted to yield reasonable outcomes, such as "pedestrians should remain on the sidewalk", "vehicles should follow lane centerlines", etc. \cite{traffic-generation-8569938, traffic-generation-prakash2020structured, traffic-generation-yang1996microscopic}. However, these heuristics are not capable of capturing complex behaviors found in real-world traffic scenarios.

\begin{figure}
    \captionsetup{font=scriptsize}
    \centering
    \includegraphics[clip, trim=0.0cm 4.1cm 10.0cm 3.0cm, width=1.0, width=1\linewidth]{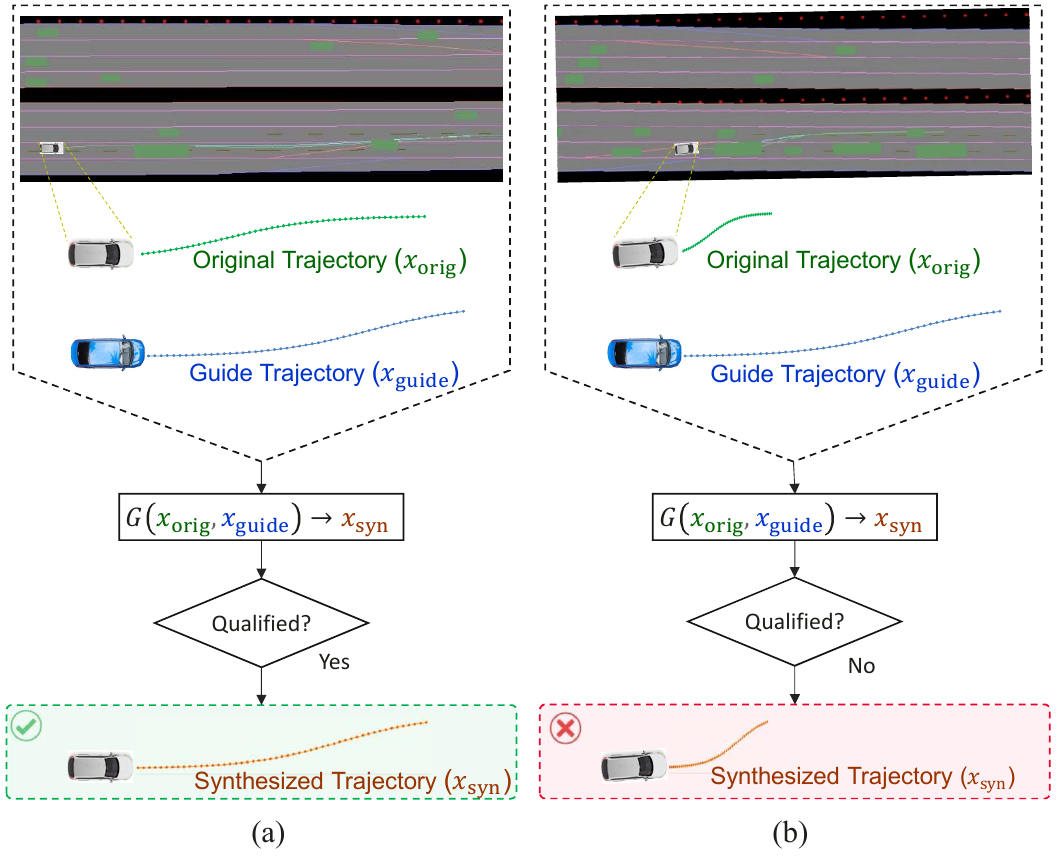}
    \caption{Synthesis steps for generating a new trajectory using our defined transformative process applied to two trajectories within each pair: (a) one satisfying predefined acceptance criteria for augmentation, and (b) one failing to meet the specified acceptance criteria for augmentation.}
    \label{fig:our-proposed-approach-diagram}
\end{figure}

Other common approaches, such as \cite{SMOGN-pmlr-v74-branco17a}, involve using methods based on SMOTE \cite{SMOTE}, where oversampling is applied to the minority class while under-sampling is performed on the majority class. However, this naive technique falls short in creating diverse driving scenarios, limiting its potential for enhancing model generalization. Moving beyond conventional methodologies, the work presented in \cite{chauffeurnet} introduces a novel strategy employing a random synthesizer to perturb demonstration data trajectories. The midpoint of these trajectories undergoes perturbations based on a uniform distribution, while the start and end points remain unchanged. Although their approach exhibited an enhancement in model performance, the use of random perturbations in generating augmented trajectories introduces the potential for deviation from expert behavior. Several other attempts have been made to employ learning-based methods for generating new trajectories using pre-trained generative models \cite{feedback_synthesizer, CMTS, Nachiket_Convolutional_Social_Pooling_2019, Ilya_seq2seq_2014, imitation-is-not-enough-lu2023imitation}. For example, in \cite{feedback_synthesizer}, the authors introduced a feedback synthesizer to generate new synthetic trajectories and augment the training dataset. However, this dependency raises the concern that any distributional shift in the model's behavior may be transferred to the synthesized scenarios, potentially leading to the creation of non-realistic trajectories or introducing bias into the dataset.  

To address the aforementioned limitations, we propose a framework to leverage demonstration data to create synthetic trajectories that maintain the essential characteristics of the original expert data. To accomplish this objective, we first train a model to cluster trajectories, identifying minority yet safety-critical scenarios. Presuming that trajectories within the same cluster exhibit similar characteristics, we will utilize trajectories from the identical cluster to generate new trajectories through a geometric transformation. To ensure feasibility of the new trajectories, we check their quality before adding them to the training dataset. Our main contributions in this work are:
\begin{itemize}
    \item We establish a trajectory clustering method based on autoencoders (AEs) to extract challenging driving scenarios.
    \item We develop a data augmentation framework to generate trajectories in extracted safety-critical but minority driving scenarios while preserving similarity to expert data.
    \item We validate the effectiveness of our data augmentation algorithm in urban and highway driving scenarios, namely InD \cite{InD} and TrafficJams, respectively. Experiment results exhibit that our approach enhances the closed-loop performance of data driven planning models, which contributes to mitigation of distributional shift.
\end{itemize}
An overview of our methodology is illustrated in Fig. \ref{fig:our-proposed-approach-diagram}.

\section{Methodology}
The key aspect of our research lies in the augmentation of expert driving trajectory data—a beneficial endeavor within the realm of autonomous planning. The primary challenge we confront is the synthesis of trajectories that not only deviate meaningfully from the training data but also uphold the intrinsic characteristics of the clusters to which they are assigned. Furthermore, the augmented trajectories must adhere closely to established rules and regulations, ensuring the preservation of safety criticality inherent in the original training data. In this paper, we delve into the specifics of our trajectory augmentation proposal and its integration with clustering methods. Subsequent sections will provide an overview of our methodology. 

\subsection{Step-1: Scenario Clustering} 

\begin{figure}
    \captionsetup{font=scriptsize}
    \centering
    \includegraphics[clip, trim=0.5cm 10.0cm 16.0cm 3.5cm, width=0.75\linewidth]{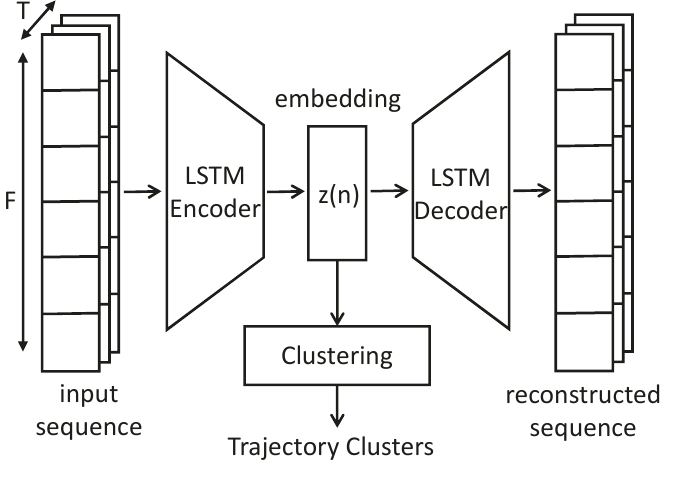}
    \caption{Architecture of our LSTM-based autoencoder.}
    \label{fig:LSTM-AE}
\end{figure}

In the pursuit of an effective augmentation method, it is imperative to appropriately cluster data samples into distinct groups. The challenge arises due to the high dimensionality of input samples, rendering direct clustering in the input space insufficient for capturing underlying patterns and structures. To address this, we leverage a compact and meaningful representation of input samples, and apply the clustering algorithm in the embedding space. As depicted in Fig. \ref{fig:LSTM-AE}, we employ AEs to derive latent representations of the data samples. Given that our samples represent segments of vehicle trajectories, it becomes essential to consider temporal dependencies between different segments. To tackle this challenge, we integrate Long Short-Term Memory (LSTM) units into the structure of our AE's encoder and decoder. The encoder's hidden state, $z(n)$, serves as the embedding representation, and the loss is defined as the Mean Squared Error (MSE) between the input sample and the generated one. The dimensions of our input data are denoted by $F \times T$, where $F$ and $T$ represent the feature size and time horizon of the trajectory samples, respectively. The input features specifically comprise the vehicle's speed, acceleration, heading, and rate of turn. Once the latent space representation is acquired, we apply K-means clustering directly to the embeddings for effective data grouping \cite{Unsupervised_Driving_Event_Discovery_Kreutz_2022}.

\subsection{Step-2: Trajectory Synthesizing} 
The initiation of our trajectory synthesis process involves pairing two candidate trajectories within a single cluster. Our working assumption is rooted in the premise that trajectories belonging to the same cluster inherently share analogous characteristics. This similarity extends to aspects such as the manner in which an expert driver executes a lane change or decelerates in response to a sudden stop from the lead vehicle. Consequently, we embark on the synthesis of a new trajectory through a defined transformative process applied to the two trajectories within each pair. For clarity, each pair is characterized by:
\begin{itemize}
    \item \textbf{Original Trajectory}: The trajectory whose start and end points serve as the basis for generating the new trajectory.
    \item \textbf{Guide Trajectory}: The trajectory whose shape is utilized as a guiding reference for matching while establishing connections between the start and end points of the original trajectory.
\end{itemize}
First, we modify the guide trajectory to align its start and end points with those of the original trajectory. 
This alignment is crucial because, after the scenario trajectory concludes, the ego must seamlessly match the remaining waypoints in the training dataset. As a part of the transformation, we adjust the waypoints of the newly synthesized trajectory through interpolation to align with the number of points in the original trajectory.
\begin{equation}
\begin{cases}
    \mathbf{x}' = G \left( \mathbf{x_{\mathrm{original}}}, \mathbf{x_{\mathrm{guide}}} \right) \\
    x'|_{i=0}=x_{\mathrm{original}}|_{i=0} \\
    x'|_{i=N}=x_{\mathrm{original}}|_{i=N}
\end{cases}
\end{equation}\label{eq: transform-trajectory-01}
where $\mathbf{x_{\mathrm{original}}}=\left\{x_{\mathrm{original}}\right\}_{i=1}^{N}$, $\mathbf{x_{\mathrm{guide}}}=\left\{x_{\mathrm{guide}}\right\}_{j=1}^{M}$, and $\mathbf{x'}=\left\{x'\right\}_{i=1}^{N}$, representing the sets of original, guide, and new trajectory points respectively, and $G$ is the geometric transformation function, mapping $\mathbf{x}$ to $\mathbf{x}'$.

In our present setup, we opt for the simplest transformation approach for the sake of simplicity, employing a basic function involving rotation followed by scaling, satisfying the same boundary conditions as Eq. \ref{eq: transform-trajectory-01}.
\begin{equation}
\mathbf{x}'=G \left(\mathbf{x_{\mathrm{guide}}} \right)=\mathbf{A}\left( \mathbf{R}\mathbf{x_{\mathrm{guide}}} \right)+\mathbf{B}
\end{equation}\label{eq: transform-trajectory-02}
The primary advantage of employing these transformation functions lies in their ability to maintain the shape of the guide trajectory. This facilitates the transfer of the guide trajectory's characteristics into the original trajectory, essentially causing the original trajectory to adopt the shape of the guide trajectory. Fig. \ref{fig:our-proposed-approach-diagram} demonstrates an example of synthesizing a new trajectory through abovementioned steps. This iterative process is then systematically repeated until all conceivable pairs within each cluster have been exhaustively explored. 
It is important to highlight that the potential for transformation functions extends beyond our current method. More complex transformations, such as leveraging a deep learning-based model with a suitable loss function, are untapped possibilities. We postpone the exploration of these advanced techniques to future research efforts.

\subsection{Step-3: Quality Assurance Checks} 
During the generation of a new trajectory, there is a possibility that some of the generated trajectories might violate fundamental traffic rules. To avoid incorporating such trajectories into the training dataset, we implement rigorous quality checks to guarantee that the generated trajectories satisfy the minimum criteria for quality. In the current implementation, the criteria include:

 \begin{enumerate}
     \item \textbf{Trajectory Similarity Checks:} 
        Both originator trajectories must satisfy similarity criteria to serve as the groundwork for generating new trajectories. In this context, we posit that these criteria act as a mechanism for validating whether the context surrounding the ego shares analogous characteristics. In our present implementation, this validation is accomplished by calculating longitudinal and lateral offsets in the Frenet frame, along with the mean velocity error between the two originator trajectories.
        \begin{equation}
        \begin{split}
            \delta_{\mathrm{Lon}} = \norm{s_{\mathrm{guide}} - s_{\mathrm{original}}} < \Delta_{\mathrm{Lon}} \\
            \delta_{\mathrm{Lat}} = \norm{d_{\mathrm{guide}} - d_{\mathrm{original}}} < \Delta_{\mathrm{Lat}} \\
            \delta_{\mathrm{Vel}} = \norm{v_{\mathrm{guide}} - v_{\mathrm{original}}} < \Delta_{\mathrm{Vel}} \\
        \end{split}
        \end{equation}\label{eq: acceptance-criteria}
        where $\Delta_{\mathrm{Lon}}$ and $\Delta_{\mathrm{Lat}}$ are longitudinal, lateral offset thresholds and $\Delta_{\mathrm{Vel}}$ is maximum average velocity error threshold. 

     \item \textbf{Trajectory Quality Checks:}
        To further ensure the synthesized trajectories meet fundamental quality standards, a multi-faceted quality check is employed. This includes scrutiny for:
        \begin{itemize}
            \item Safety Compliance (e.g., collision-free status, avoidance of near-miss trajectories, etc.).
            \item Comfort Metrics (max acceleration, jerk, curvature, etc.). 
            \item Adherence to Traffic Rules. 
            \item Trajectory finish time or average speed / acceleration in the planned trajectory. 
        \end{itemize}
        As of the current implementation, collision checks have been specifically integrated into this quality assessment step through examining the waypoints of the new trajectory against the waypoints of all dynamic agents present in the scene at each time step.
\end{enumerate}

\begin{figure}[t]
    \captionsetup{font=scriptsize}
    \centering
    \includegraphics[width=\linewidth]{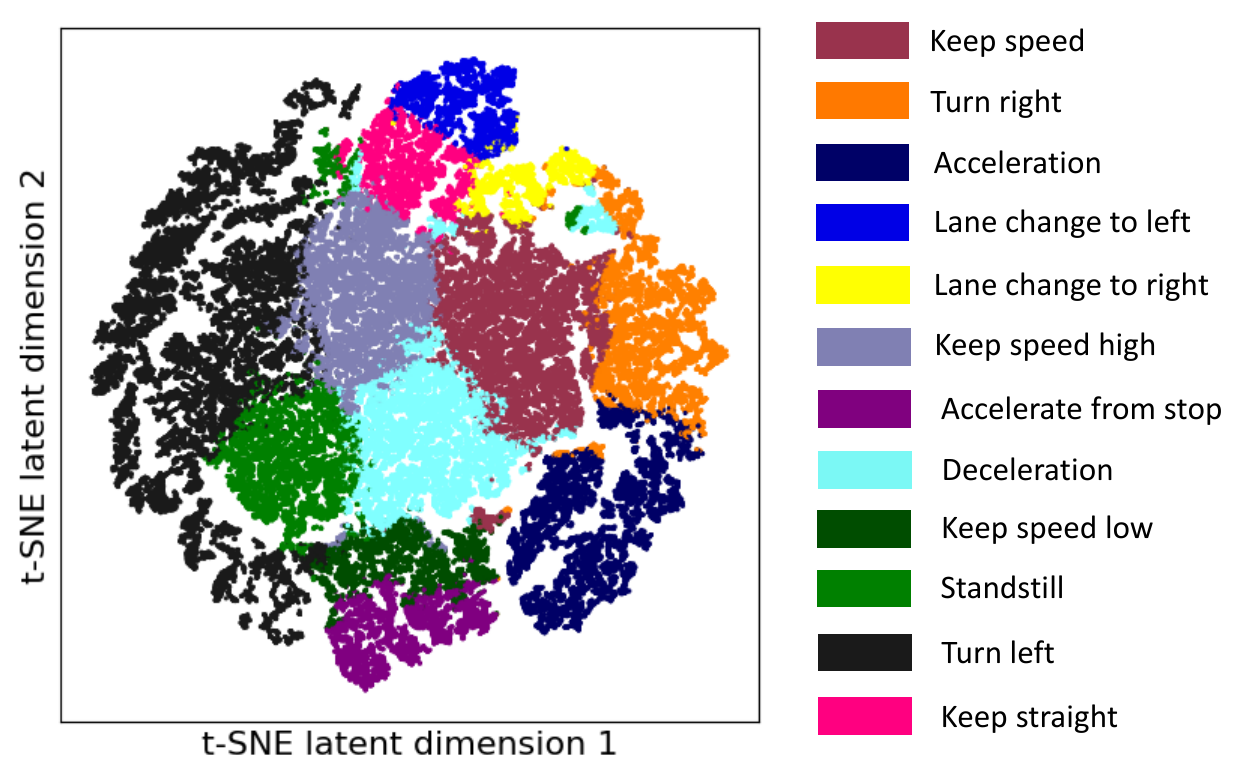}
    \caption{t-SNE of the embeddings and their corresponding clusters.} 
    \label{tsne}
\end{figure}

Only when a synthesized trajectory meets both criteria is it deemed eligible for inclusion in the training dataset. In summary, employing our approach enables the generation of a maximum of $n_{\text{cl}}\left(n_{\text{cl}}-1\right)$ new trajectories, depending upon the number of trajectories meeting the specified criteria outlined above. Here, $n_{\text{cl}}$ corresponds to the total number of original trajectories within a given cluster. 
\section{Results and Discussions}
To assess the efficacy of our approach, we evaluate our methodology on two distinct datasets: the InD dataset \cite{InD}, which comprises urban driving data around intersections, and the TrafficJams dataset, which focuses on highway driving scenarios with congestions. Both datasets, InD and TrafficJams, are novel collections of naturalistic vehicle trajectories recorded at German intersections and highways, respectively. They encompass the most compelling scenarios encountered at intersections and during the merging and exiting of highways affected by congestion.

\subsection{Scenario Clustering and Trajectory Synthesizing}
As our methodology operates on clusters of expert trajectory data, a crucial aspect involves extracting meaningful trajectory groups. Our input dataset is expressed as
\begin{equation}
    \mathcal{D}=\{(\mathbf{v}_{t:t+T}, \mathbf{\Dot{v}}_{t:t+T}, \mathbf{h}_{t:t+T},\mathbf{\Dot{h}}_{t:t+T})\},  
\end{equation}
where $\mathbf{v}$, $\mathbf{\Dot{v}}$, $\mathbf{h}$, and $\mathbf{\Dot{h}}$ are sequence of speed, acceleration, heading, and rate of turn between time $t$ and $t+T$, respectively. In our experiments, we use $T=3$s with a frame rate of $10$Hz.

\begin{figure}[t]
    \captionsetup{font=scriptsize}
    \centering
    \subfloat[]
    {
        \includegraphics[clip, trim=0.2cm 14.9cm 15.4cm 3.0cm, width=1.0\linewidth]{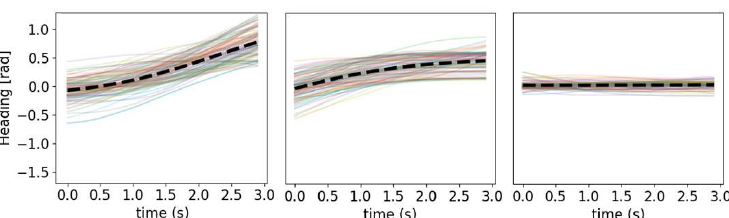}%
    }
    \\
    \subfloat[]
    {
        \includegraphics[clip, trim=0.2cm 14.9cm 15.4cm 3.0cm, width=1.0\linewidth]{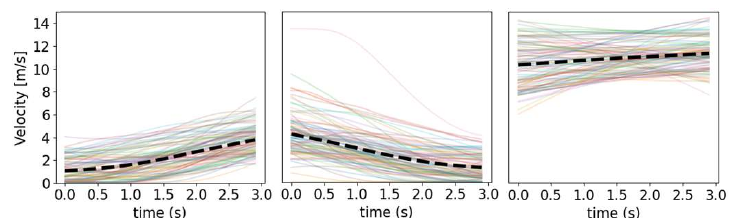}%
    }
    \caption{Examples of the results of our trajectory clustering approach. (a) Sample clustered headings, representing approaching intersection for turn left (left), leaving intersection after turn left (center), and keeping lane (right). (b) Examples of clustered velocities, representing acceleration (left), deceleration (center), and monotonous velocity (right).
    }
    \label{fig:trajectory-profile}
\end{figure}

Fig. \ref{tsne} represents the t-SNE of the embeddings and their corresponding scenarios. Also, Fig. \ref{fig:trajectory-profile} shows examples of the trajectory profiles in some of these clusters. While our clustering method can derive an arbitrary number of clusters for trajectories with specific durations, it is essential to ensure that the grouped trajectories exhibit shared characteristics, a crucial prerequisite for augmentation. For instance, if two trajectories—one involving acceleration and another deceleration—happen to be grouped together, our subsequent quality check would reject them for augmentation. In essence, achieving improved clustering results is imperative to enhance the likelihood of meeting the aforementioned requirements. Therefore, we refined the clustering hyperparameters to assign groups to the most similar trajectories and manually inspected and combined clusters that showed similar behavior, thereby increasing the chances of satisfying the specified criteria. This led to the consolidation of trajectories into the subsequent cluster within our urban driving dataset:
\begin{itemize}
    \item Approaching the intersection for turning left or right.
    \item Leaving the intersection after turning left or right.
    \item Stopping at the intersection and starting after a stop at the intersection. 
\end{itemize}

\begin{figure}[t]
    \captionsetup{font=scriptsize}
    \centering
    \includegraphics[width=1\linewidth]{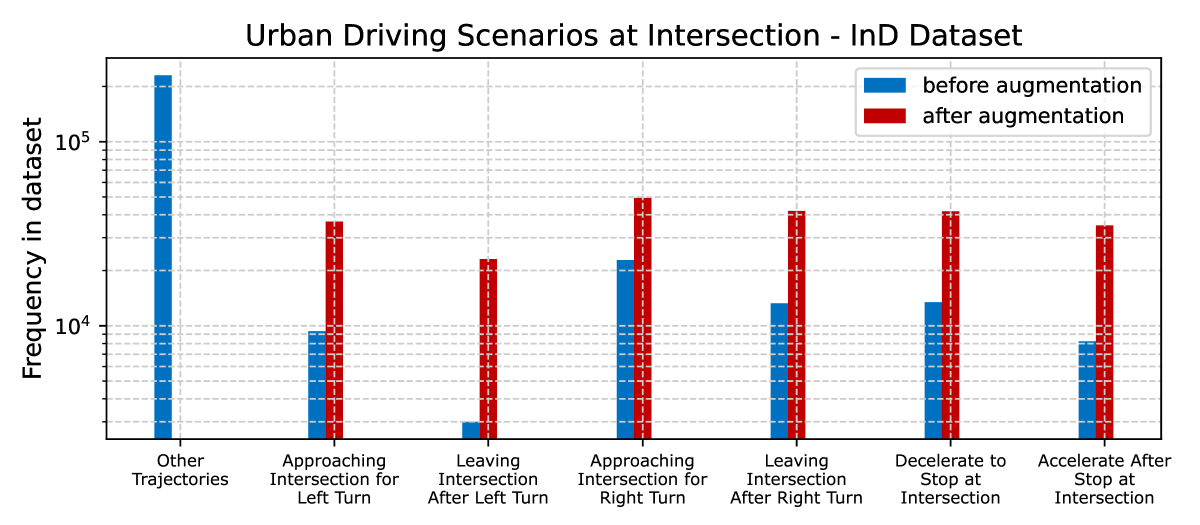}
    \includegraphics[width=1\linewidth]{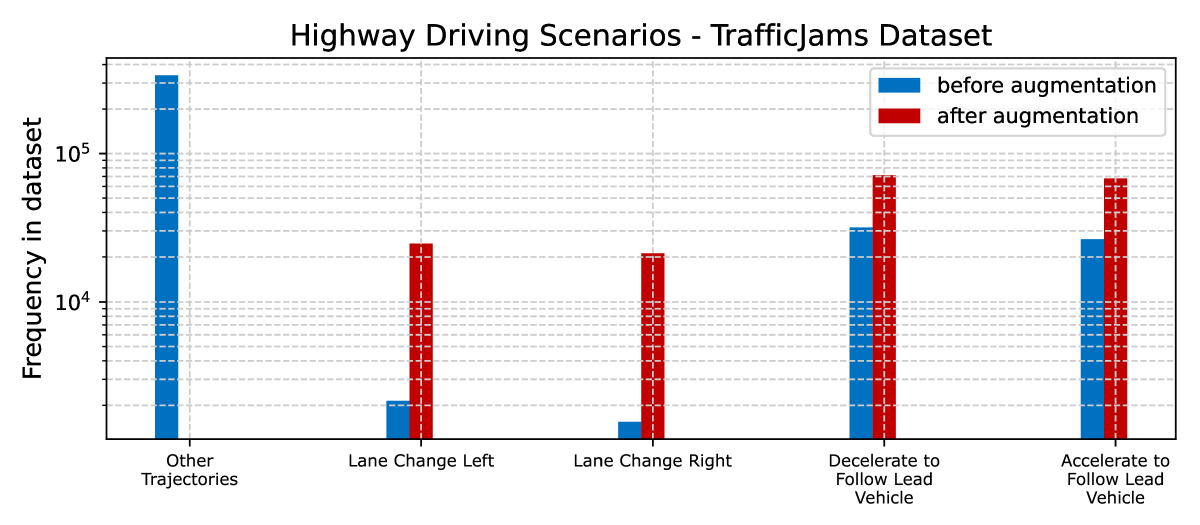}
    \vspace{-7mm}
    \caption{Comparison of data cluster percentages before and after augmentation using our approach, for our urban driving scenarios (top) and highway driving scenarios (bottom).}
    \label{fig:data-distributions}
\end{figure}

Applying a similar approach, the highway driving scenarios are classified into the subsequent clusters:
\begin{itemize}
    \item Lane changes, both left and right.
    \item Acceleration and deceleration to align the ego's actions with the traffic flow while maintaining the lane.
\end{itemize}
Other clusters that did not fit into the above categories were combined into a single "other trajectories" cluster.
It is important to note that, based on our offline investigations, a significant proportion of failures in both our highway and urban driving scenarios originate from the trajectory data associated with these clusters. Hence, we posit that these clusters effectively represent safety-critical scenarios to a considerable extent. Fig. \ref{fig:data-distributions} illustrates the comparison of data cluster frequencies before and after augmentation through our approach. The updated frequencies for each cluster are determined by the ratio of the square root of the original data frequencies within each cluster \cite{huang2016learning,cui2019class}.

\subsection{Closed-Loop Results}
We evaluate our proposed methodology within an imitation learning framework for planning, adapted from TNT \cite{Zhao_TNT}, which is a motion forecasting algorithm. We have devised a tailored version of the TNT pipeline, incorporating dynamic states of the present agents via bicycle filter, to make it suitable for motion planning purposes to enhance its closed-loop performance.  
InD dataset comprises slightly over 300k samples for training. The TrafficJams dataset encompasses over 1.5M training samples in total. However, given the perceived similarity among several data points, we could randomly select 400k samples for training without compromising generality.

In the highway driving scenario, we deliberately selected the most challenging scenarios from the test dataset through manual labeling. These scenarios were characterized by presenting congested traffic jams, where the ego vehicle had to decelerate or fully stop behind the lead vehicle and then accelerate to rejoin the traffic flow. Likewise, within our urban driving dataset, we identified scenarios from the test dataset, believed to pose greater learning challenges for our imitation learning framework when relying solely on the original training data. These scenarios were chosen in a manner that demands the ego vehicle to execute more intricate maneuvers to successfully navigate the scenario.  
Our closed-loop simulation is based on a non-reactive simulation replaying the recorded waypoints of other agents in the datasets.
The closed-loop results are evaluated through the following metrics:
\begin{itemize}
    \item \textbf{Progress Percentage}: This represents the ratio of the driven distance for a scenario in the closed-loop to that of the ground truth.
    \item \textbf{Success Percentage}: Indicates the proportion of scenarios that concluded successfully.
    \item \textbf{Collision Percentage}: Reflects the percentage of scenarios that resulted in a collision between the ego and other dynamic agents.
    \item \textbf{Mean Distance Between Collisions (MDBC)}: This metric denotes the average driven distance of the ego before encountering collisions with other dynamic agents.
\end{itemize}
Our experiments exhibited that incorporating the augmented safety-critical trajectories into the training dataset improved closed-loop performance in both our urban and highway driving scenarios. 
We observed that our cluster-based trajectory augmentation method demonstrated improvements across all mean values of the closed-loop metrics in both our urban and highway driving scenarios, as presented in Tables \ref{tab:closed-loop-traffic-jams} and \ref{tab:closed-loop-ind}. It is worth mentioning that our approach contributed to enhanced robustness in imitation learning, evidenced by reduced ratio of standard deviation over mean (i.e. coefficient of variation) of progress percentage among different runs with distinct seeds.

\begin{table}[t]
    \captionsetup{font=scriptsize}
    \centering
    \setlength{\tabcolsep}{2pt}
    \renewcommand{\arraystretch}{1.5}
    \caption{Impact of integrating our trajectory augmentation method on closed-loop performance within our highway driving scenarios (TrafficJams dataset). The outcomes are derived by averaging the results from three training runs of our model, each conducted with a distinct seed.}
    \begin{tabular}{
      >{\centering}p{0.1\textwidth}
      >{\centering}p{0.0\textwidth}
      >{\centering}p{0.08\textwidth}
      >{\centering}p{0.08\textwidth}
      >{\centering}p{0.08\textwidth}
      >{\centering\arraybackslash}p{0.10\textwidth}
    }
        \toprule
        \textbf{Augmented Clusters} & & \textbf{progress \%} & \textbf{success \%} & \textbf{collision \%} & \textbf{mdbc [m]} \\
        \hline
        \multirow{1}{6em}{N/A} &  & $58.5\pm5.2$ & $23.4\pm5.2$ & $75.4\pm5.4$ & $277.7\pm44.6$ \\
        \hline
        \multirow{1}{6em}{LCL+LCR} &  & $59.2\pm4.1$ & $26.4\pm4.2$ & $71.9\pm4.0$ & $294.1\pm35.0$ \\
        \hline
        \multirow{1}{6em}{Dec.+Acc.} &  & $61.3\pm4.2$ & $25.3\pm4.2$ & $74.0\pm4.8$ & $289.4\pm41.5$ \\
        \hline
        \multirow{2}{6em}{Dec.+Acc.+ LCL+LCR} &  & $64.5\pm3.8$ & $31.0\pm4.8$ & $68.0\pm5.3$ & $337.3\pm51.6$ \\
        &  &  &  &  &  \\
        \hline
        \multirow{2}{6em}{Dec.+Acc.+ LCL+LCR upsample \cite{SMOTE}} &  & $39.8\pm5.7$ & $12.1\pm4.0$ & $86.5\pm4.1$ & $159.0\pm25.1$ \\
        &  &  &  &  &  \\
        \bottomrule
    \end{tabular}
    \label{tab:closed-loop-traffic-jams}
\end{table}

\begin{table}[t]
    \captionsetup{font=scriptsize}
    \centering
    \setlength{\tabcolsep}{2pt}
    \renewcommand{\arraystretch}{1.5}
    \caption{Impact of integrating our trajectory augmentation method on closed-loop performance within our urban driving scenarios (InD dataset). The outcomes are derived by averaging the results from three training runs of our model, each conducted with a distinct seed.}
    \begin{tabular}{
      >{\centering}p{0.10\textwidth}
      >{\centering}p{0.00\textwidth}
      >{\centering}p{0.08\textwidth}
      >{\centering}p{0.08\textwidth}
      >{\centering}p{0.08\textwidth}
      >{\centering\arraybackslash}p{0.10\textwidth}
    }
        \toprule
        \textbf{Augmented Clusters} &  & \textbf{progress \%} & \textbf{success \%} & \textbf{collision \%} & \textbf{mdbc [m]} \\
        \hline
        \multirow{1}{8em}{N/A} &  & $83.3\pm1.6$ & $54.7\pm5.4$ & $40.3\pm6.1$ & $137.5\pm26.2$ \\
        \hline
        \multirow{1}{8em}{Turns} &  & $87.5\pm0.2$ & $61.7\pm2.1$ & $35.0\pm2.4$ & $156.7\pm14.0$ \\
        \hline
        \multirow{1}{8em}{Dec. + Acc.} &  & $87.8\pm1.8$ & $59.7\pm1.2$ & $35.7\pm1.2$ & $153.7\pm3.7$ \\
        \hline
        \multirow{1}{8em}{Dec.+Acc.+Turns} &  & $88.7\pm1.1$ & $61.3\pm4.2$ & $36.7\pm4.0$ & $151.2\pm20.8$ \\
        \bottomrule
    \end{tabular}
    \label{tab:closed-loop-ind}
\end{table}

\begin{figure}[t]
    \captionsetup{font=scriptsize}
    \centering
    \subfloat[\scriptsize]
    {
        \includegraphics[clip, trim=0.0cm 11.0cm 8.0cm 3.0cm, width=1.0\linewidth]{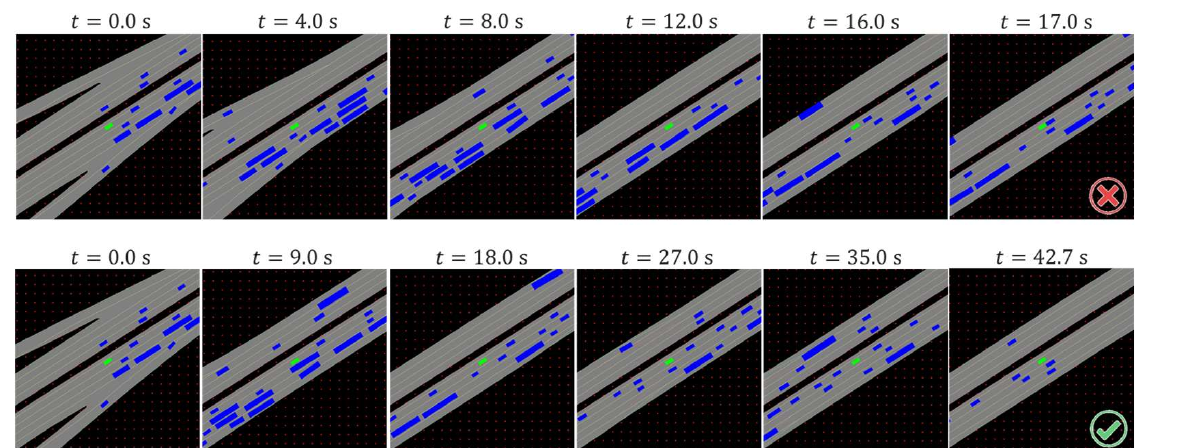}%
    }\\
    \subfloat[\scriptsize]
    {
        \includegraphics[clip, trim=0.0cm 11.0cm 8.0cm 3.0cm, width=1.0\linewidth]{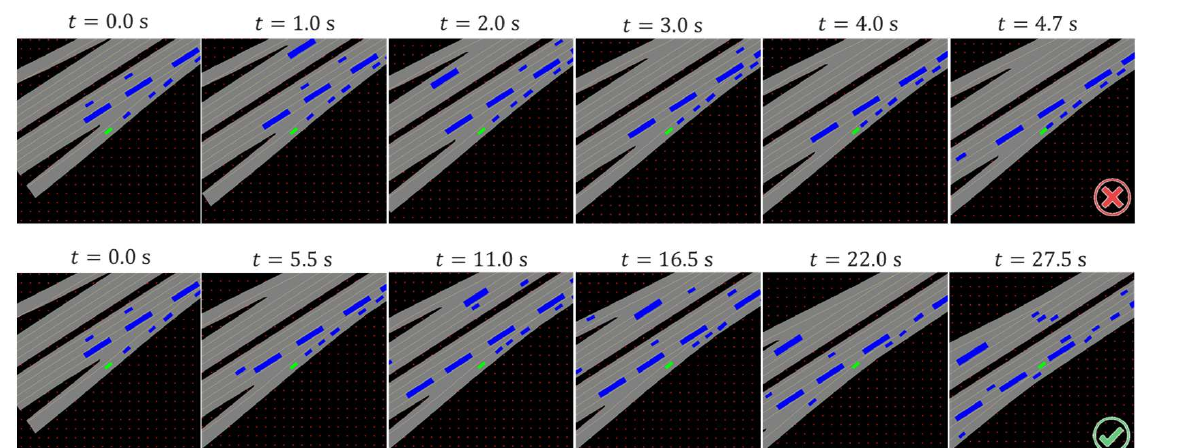}%
    }
    \caption{Sample snapshots of the closed-loop behaviors, illustrating cases with and without trajectory augmentation. The top and bottom images for each case are corresponding to scenarios without and with augmentation, respectively.}
    \label{fig:closed_loop_performance_snapshots}
\end{figure}

In addition, the behavior of the ego vehicle exhibits overall improvement, evident not only in quantitative metrics but also in qualitative aspects. Fig. \ref{fig:closed_loop_performance_snapshots} presents two sample snapshots depicting closed-loop behaviors—one with trajectory augmentation and one without. In the first scenario, ego vehicle fails to adapt its acceleration to the traffic flow, resulting in a collision with the lead vehicle (Fig. \ref{fig:closed_loop_performance_snapshots}-a, top). Conversely, training with augmentation using our approach enhances this behavior, allowing the ego vehicle to persist significantly longer without collision (Fig. \ref{fig:closed_loop_performance_snapshots}-a, bottom). In the second scenario, the ego vehicle attempts to merge into the traffic coming from the left lane of the highway, where the merging lanes experience congestion. In the training outcomes without augmentation, the ego vehicle encounters challenges in adjusting its maneuvers to the merging traffic, resulting in a collision with the lead vehicle (Fig. \ref{fig:closed_loop_performance_snapshots}-b, top). In contrast, by employing our approach, the ego vehicle adeptly executes lane changes, seamlessly merging into the incoming lane (Fig. \ref{fig:closed_loop_performance_snapshots}-b, bottom). It is noteworthy that we did not explicitly augment merging scenarios in our current implementation. However, our approach proved beneficial in enhancing the generalization of closed-loop results, extending its efficacy to scenarios not specifically targeted for augmentation \footnote{Additional examples have been included in a supplementary video, accessible through \href{https://youtu.be/ASY-OZ93if8}{https://youtu.be/ASY-OZ93if8}}.

Furthermore, to gauge the efficacy of our efforts in enhancing closed-loop performance, we conducted an additional experiment within the highway scenario, employing up-sampling technique for the aforementioned minority clusters \cite{SMOTE}. Following the conclusion of this experiment (see Table \ref{tab:closed-loop-traffic-jams}), it became apparent that our methodology exhibited a more pronounced effectiveness in improving results compared to the up-sampling approach.
\section{Conclusion}
In this study, we introduced a trajectory augmentation method designed to preserve the fundamental characteristics of the original expert trajectories. Our experiments exhibit that training an imitation learning model using these augmented trajectories can significantly improve closed-loop performance. Additionally, we highlighted that our method not only yields improved quantitative closed-loop metrics but also results in enhanced qualitative behavior, thereby improving the generalization of our imitation learning model. Our evaluation spanned two distinct trajectory datasets (urban and highway driving scenarios), illustrating the generalization capacity of our approach in enhancing robustness for imitation learning. For future work, we would like to explore various implementations of the geometric transformation function $G$ in Eq. \ref{eq: transform-trajectory-01} and possibility of employing a learning-based approach to generate the trajectories similar to expert data. 

\bibliographystyle{IEEEtran.bst}

\typeout{}
\bibliography{01-references.bib}

\end{document}